# Competency Model Approach to AI Literacy:
# Research-based Path from Initial Framework to Model

**Farhana Faruqe, Ryan Watkins, and Larry Medsker**
Human-Technology Collaboration Lab
The George Washington University, Washington, DC
faruqe@gwu.edu, rwatkins@gwu.edu, lrm@gwu.edu

**Abstract**

The recent developments in Artificial Intelligence (AI) technologies challenge educators and educational institutions to respond with curriculum and resources that prepare students of all ages with the foundational knowledge and skills for success in the AI workplace. Research on AI Literacy could lead to an effective and practical platform for developing these skills. We propose and advocate for a pathway for developing AI Literacy as a pragmatic and useful tool for AI education. Such a discipline requires moving beyond a conceptual framework to a multi-level competency model with associated competency assessments. This approach to an AI Literacy could guide future development of instructional content as we prepare a range of groups (i.e., consumers, coworkers, collaborators, and creators). We propose here a research matrix as an initial step in the development of a roadmap for AI Literacy research, which requires a systematic and coordinated effort with the support of publication outlets and research funding, to expand the areas of competency and assessments.

## 1 Introduction

AI technologies are impacting individuals and our society at an increasing rate and in a number of ways. The benefits of AI applications are amazing, while our lived experience is revealing serious risks and the potential catastrophic failures in our complex and interdependent systems. These cases, and revelations in the media about bias in routine applications of AI, create doubts among the public about the trustworthiness of products we have otherwise taken for granted. In our times of widespread misunderstanding and misinformation, AI Literacy is emerging as an essential factor that will determine the future of AI. Building a strong foundation for this field (including associated educational and training assets) is necessary for understanding the real benefits and risks of AI. Science-based research is therefore equally necessary to identify the elements of, and establish trust in, the field of AI Literacy. This paper proposes a structured approach, with a focus on competencies and metrics for measuring levels required at different knowledge levels, to survive and thrive in the new world of AI.

In this paper, a research matrix is proposed as the initial step in developing a roadmap for AI Literacy research. A major recommendation is that researchers use unique, different behavioral anchors to categorize the roles of AI users (consumer, coworker, collaborator, creator). Establishing AI literacy will be difficult and complex to prepare foundational AI knowledge truly needed in those roles. The results given in this paper, instructive and not a complete competency model, are based on our summary of various approaches to creating full models for other literacies (e.g., digital and media) combined with our experience in applied AI. We suggest a relationship between the items in the model to the framework.

The levels of AI Literacy (and related AI ethics) required in different groups within our society vary across a spectrum -- from AI researchers and developers to workers and consumers. As a result, AI Literacy is also essential for policymakers as they make important decisions across this spectrum so that laws and regulations have the desired effects and individuals and societal groups are protected. Though various groups in the continuum can be considered, the focus for this paper comprises the (i) consumer as someone who uses the outputs of AI to improve their work or life, (ii) co-worker as someone knows the basics of how the AI systems work and uses AI outputs in the work, (iii) collaborator as someone who works alongside one or more AI systems to improve each others performance, and (iv) creator as someone who develops and tests new AI systems and underlying models from the ground up (see Figure 1). Each of the groups thereby requires a different level of AI Literacy knowledge and skills.

The unknown aspects of work in the future present many challenges to educators and educational institutions at all levels, especially in terms of the multiple and diverse roles that AI technologies will likely play in all sectors of the economy. From AI agents who are our co-workers to the deployment of AI systems in education that help prepare learners for the workplace, the specific implications of AI in all aspects of learning (from K12 preparation through ongoing professional development) are still largely unknown. Nevertheless, the important role of AI technologies has become clear [OECD.AI Policy Observatory, 2021; Yeo, 2020; Harrod, 2020] and more so the necessary capacity-building of people to collaborate and work alongside AI in the future.

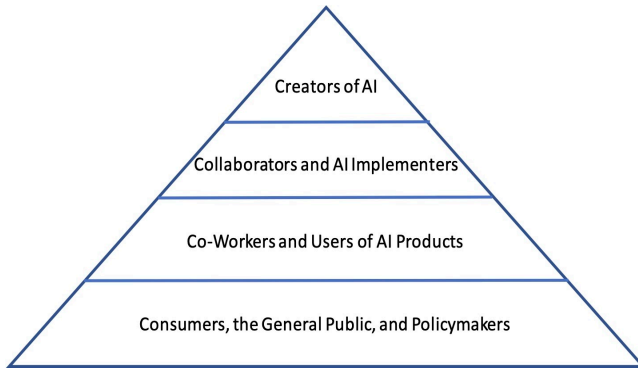

Figure 1: Focused Group

The increasing importance of accelerating the rate at which people (within diverse jobs and with varied experiences) can gain new skills or knowledge required for positions in an AI-infused workforce has been in the spotlight for the last several years [McKinsey Report, 2018]. For some workers this will require transitions to new careers, whereas for others this may require acquiring increasing technical knowledge in order to work alongside or collaborate with AI technologies in the workplace. For most it will require, at minimum, the capacity to appropriately use the outputs of AI systems in their decision-making. That is, AI Literacy at varying levels is necessary to achieve career transition pathways for workers who lose jobs due to AI and other automation technologies.

Formal and informal educational opportunities for workers transitioning to new jobs in the AI economy involve two areas of skills and knowledge: (i) knowledge of AI and how to use AI systems in hybrid (i.e., human-machine) teams and (ii) non-AI learning necessary for new automation-related spin-off jobs predicted in the new AI economy. In each case, workers must re-skill quickly to learn content and build skills for team work (including teams with AI collaborators). This type of training, on a large scale, is however not a good match for course- and degree-oriented traditional education institutions. Massively open online systems could be a critical element of the solutions, but an important policy-related issue is who will provide those systems (private sector, public sector, non-profits, etc.). In order to make these decisions, however, many of the technology issues for the enormity of the need will require scaled up systems involving teams of humans (e.g., instructors, facilitators), cognitive assistants (e.g., automated learning assistants), and AI-based learning management (e.g., individualized learning paths) to make learning systems manageable at a very large scale.

Predictive career transition models are also necessary for workers seeking new careers after losing jobs due to automation through AI and other technologies, including a focus on issues of social inequality for different careers and demographics. This includes many challenges for many demographics including, for example, the plight of older workers experiencing employment disruption and having difficulty shifting to new careers before and beyond retirement age. These requirements for improving labor markets, forming workforce development policies, and creating new career pathways require extensive and careful analysis of what levels of knowledge of AI workers require and what kinds of specific tasks they must be able to do. To have valuable influence, AI Literacy researchers must define pragmatic groupings in the new AI society and the levels of AI knowledge/skills required in each group.

The introduction of new knowledge, skills, attitudes, and abilities for working alongside new AI technologies (many of which haven't been invented yet) is a unique aspect of the challenges for educators and educational institutions. There is however emerging literature and a new focus on AI Literacy [Long and Magerko, 2020] that is beginning to lay the foundations on which curricula, courses, certifications, learning programs, MOOCs, and other educational tools for AI Literacy can be built (e.g., Payne, 2021). The current proposals for AI Literacy frameworks are however just that, early proposals and are not mature frameworks that can be utilized from the classroom and recruitment fairs, to hiring decisions and promotion criteria. To serve the emerging workplace of the future, we posit that systematic and coordinated efforts, including a robust research initiative, is required to quickly (i.e., within 3 to 5 years) affirm the utility of AI literacy for educational and related applications.

## 2 AI Literacy

Over the last 30 years there has been no shortage of new "literacies" proposed as guidance for educators and educational institutions. From Digital Literacy to Information Literacy to Media Literacy, the introduction of new literacy frameworks has been a practical tool for helping educators prepare students. To this mix AI Literacy has been added [Long and Magerko, 2020], and while this may initially appear to be an overuse of the "literacy" concept, we propose the evolution of technology literacies has been valuable, with its continuing development and expansion (to include AI Literacy) can provide useful frameworks for educators who may otherwise miss opportunities to prepare learners (of all ages) for the future.

To that end, in this paper we (i) highlight a framework for considering AI Literacy, (ii) push that framework toward a competency model with behavioral anchors, and then (iii) add to that a research matrix that can guide the development of AI Literacy research in a manner that will allow it to be-come an integral part of the educational experiences of students.

We contend that "literacy" frameworks (e.g., computer literacy, internet literacy, data literacy) offer educators and others valuable tools for understanding emerging technologies and shifts in our society. These frameworks help organize the multiple dimensions of these complex changes, such as the relationship to ethics, epistemological foundations, critical analysis, and relationships to other literacy frameworks. They also achieve this through conceptual and logical connections that are often recognizable and applicable to users of the framework (see Long and Magerko, 2020 supplemental video). Nevertheless, they have limited potential since they are routinely not yet formalized through rigorous research, not tied to behavior-based competency models, nor assessed with validated measures that align with the research and the behavior competencies. As a consequence, they are valuable but not reaching their potential.

Most are frameworks and ideas, but research is necessary to translate these useful starting places into competency-based systems that (a) provide specific performance standards for diverse levels of literacy, and (b) offer the foundation for competency assessment. Being able to assess the competency level of individuals is central to most practical applications of AI Literacy.

## 3 A Proposed AI Literacy Framework

In this article we are proposing a research-based path forward to mature AI Literacy into a tool that can guide educators and others in preparing people for the work of the future. Therefore we are not advocating for any specific AI Literacy framework -- though we find the following proposed framework by Long and Magerko [2020] to be acceptable. But it too should evolve as the research community learns more about AI literacy through systematic research efforts. For now, we are using it as a placeholder for future discussion and not as the exemplar that should be used in all research. In this manner, it illustrates how a community of researchers may approach AI Literacy for improving our understanding, our models, our curricula, our assessments, and our implementation of AI Literacy.

Long and Magerko [2020] synthesized literature from a variety of disciplines to propose 17 core competencies of AI literacy. For example:

- Competency 1 (Recognizing AI) - Distinguish between technological artifacts that use and do not use AI.
- Competency 2 (Understanding Intelligence) - Critically analyze and discuss features that make an entity "intelligent", including discussing differences between human, animal, and machine intelligence.
- Competency 16 (Ethics) - Identify and describe different perspectives on the key ethical issues surrounding AI (i.e. privacy, employment, misinformation, the singularity, ethical decision making, diversity, bias, transparency, accountability).
- Competency 17 (Programmability) -- Understand that agents are programmable.

### 3.1 From Framework to Competency Model

One essential first step in creating utility from an AI Literacy framework is the research-based development of a competency model with behavioral anchors. This allows for the translation of the framework into scalable assessments with evidence-based criterion for decision makers (such as HR departments) and for the validation of the framework as a tool that illustrates and predicts the value derived from achieving AI Literacy. Gonczi et al., [1993] state that "Under a competency-based assessment system, assessors make judgements, based on evidence, about whether an individual meets criteria specified in the profession's competency standards." (p.1) By associating measurable behaviors, for varied levels of mastery, with general competencies of the AI Literacy framework (see Table 1) a pragmatic competency model can be developed.

Our goal here is not to develop a full competency model for AI Literacy, since that is a research task. Rather here we illustrate what is necessary for maturing AI Literacy as a concept by providing example behavioral anchors that illustrate the role of defining performance measures for each level of mastery. More, and more detailed, behavioral anchors should emerge from the research on AI Literacy; generating a competency model that illustrates the multiple knowledge/skills expected at each level (i.e., consumer, co-worker, collaborator, creator). These behavioral anchors then allow researchers to identify the competencies most useful for clarifying and assessing people's skill levels; thereby allowing AI Literacy to become a practical tool for educators (such as, for developing curriculum), trainers (such as, for designing professional development), and for HR professionals (such as, for recruiting job candidates).

### 3.2 Research Pathways

Maturation of the AI Literacy discipline (moving from conceptual frameworks to research-based competency models and assessments) is best achieved through systematic, coordinated and deliberate efforts of many researchers, from many disciplines. The goal of a functional set of resources for AI literacy (from educational curricula to validated assessments) will require much research, and given the accelerating emergence of AI in the workplace, this foundation building work has to be done sooner rather than later -- especially before snake-oil AI literacy frameworks (i.e., without a scientific foundation) get developed and widely introduced into organizations and marketing campaigns -- with the potential of discrediting legitimate research-based AI Literacy. With that in mind, we propose a matrix of research topics that illustrate the variety of foundational science that can and should be used to create useful AI Literacy resources (see Table 2).

|  | Consumer | Co-worker | Collaborator | Creator |
| --- | --- | --- | --- | --- |
| Competency 9: ML Steps | Decides not to turn on AI recommendation engine since they know it uses private data to train the system | Explains to clients the general parameters for when the AI performs best | Recognizes potential bias in the results and examines characteristics of training data set | Improves model performance by systematically adjusting the number of neurons in the hidden layers |
| Competency 10: Human Role | Describes the value of humans-in-loop for sample AI systems | Identifies when human input is required to interpret AI system outputs | Provides human perspective as an input to the data and/or algorithms an AI system | Systematically applies human-centered design to their development processes |
| Competency 11: Data literacy | Knows to look at the source(s) of data before trusting an AI system | Routinely reviews the data sources to determine if there could be increasing bias | Conducts analysis of the data sources for potential bias | Routinely assesses the validity and value of data sources and updates AI system to adjust for changes in the data |

Table 1: Examples of behavioral anchors for an AI literacy competency model

|  | AI literacy competency framework | AI literacy assessments | AI literacy education | AI literacy policy |
| --- | --- | --- | --- | --- |
| Quasi-Experimental | External validation to determine of AI ethics competency actually changes behavior in the workplace | Comparison of alternative assessment strategies among groups of consumers and co-workers | Effectiveness of an e-learning curriculum for achieving collaborator level literacy skills | Difference-in-difference assessment of two more school district policies for AI literacy instruction |
| Meta Analysis | Systematic review of data literacy frameworks and their application in non-STEM majors | Systematic review of competency assessments across literacy initiatives | Systematic review of the effectiveness of MOOC programs teaching AI literacy skills | Systematic review of government policies regarding the application of low AI systems in education settings |
| Case Study/ Ethnography | Case study of an HR department using an AI literacy framework to create job descriptions | Case study of an organization HR department using an AI literacy competency model for recruiting | Case study of a AI literacy curriculum used for first year students at a community college | Case study of a state government attempt to pass regulations for AI use in medicine |
| Technology development and evaluation | Development and evaluation of program that HR departments to score resumes based on AI literacy skills identified | Development and evaluation of an App that allows students to self-assess their AI literacy skills | Development and evaluation of an AI literacy spaced repetition e-learning tool for high school students | Development and evaluation of NLP tool to analyze policy documents for AI literacy implications |

Table 2: Examples of Research for Maturing AI Literacy

## 4  Conclusions

The challenges presented by recent developments in AI technologies are straining educators and educational institutions to respond with curriculum and resources that can prepare students (of all ages) with the foundational knowledge and skills for success in the AI workplace. AI Literacy efforts offer a platform for developing these skills, and here we propose a pathway for the systematic maturing of AI Literacy as a pragmatic and useful tool for AI education. By moving from a conceptual framework to a multi-level competency model, and then developing associated competency assessments, AI Literacy can guide future development of instructional content as we prepare a range of groups (i.e., consumers, co-workers, collaborators, and creators). We also propose a research matrix as an initial step in the development of a roadmap for AI Literacy research. At the same time we recognize that a mature AI Literacy requires a network (or community) of researchers to expand the areas of competency and assessments. From publication outlets (such as, special issues of journals) and researching funding (such as, foundation financial support), the further development of AI Literacy will benefit from systematic and coordinated efforts.